\documentclass{article}

\usepackage[preprint]{neurips_2026}  

\usepackage[utf8]{inputenc}
\usepackage[T1]{fontenc}
\usepackage[colorlinks=true,linkcolor=blue,citecolor=blue,urlcolor=blue]{hyperref}
\usepackage{url}
\usepackage{booktabs}
\usepackage{amsfonts}
\usepackage{nicefrac}
\usepackage{microtype}
\usepackage{xcolor}
\usepackage{graphicx}
\usepackage{amsmath}
\usepackage{amssymb}
\usepackage{listings}
\usepackage{tikz}
\usepackage{algorithm}
\usepackage{algorithmic}

\title{WiCER: Wiki-memory Compile, Evaluate, Refine\\
Iterative Knowledge Compilation for LLM Wiki Systems}

 \author{
   Juan M.~Huerta \\
   Zinnia Tech Solutions \\
   600 Steamboat Road\\
   Greenwich, CT 06830, USA \\
   \texttt{juan.huerta@zinnia.com}
 }

\begin{document}

\maketitle

\begin{abstract}
The \emph{LLM Wiki} pattern---to compile and provide domain knowledge into a persistent
artifact and serve it to LLMs via KV cache inference---promises context access at sub-second
latency with zero retrieval failure. Realizing this requires
solving the \emph{compilation gap}: LLM-compilation distilling raw
documents into a wiki without catastrophically discarding critical facts.
We characterize this gap across 17 RepLiQA domains (6{,}800 questions):
we observe that full-context KV cache inference outperforms RAG on curated knowledge
(4.38 vs.\ 4.08/5, $7.3\times$ faster TTFT) but degrades below RAG at
scale due to attention dilution, and blind compilation fails entirely
(2.14--2.32 vs.\ 3.46, 53--60\% catastrophic failure rate).
To address the compilation gap, we propose \textbf{WiCER} (Wiki-memory Compile, Evaluate, Refine), an
iterative algorithm inspired by counterexample-guided abstraction
refinement (CEGAR) that closes this gap. WiCER evaluates compiled wikis
against diagnostic probes, identifies dropped facts, and forces their
preservation in subsequent compilations. One to two iterations recover
80\% of lost quality (mean 3.24 vs.\ 3.47 for raw full-context
across the 15 topics with baselines), reducing catastrophic failures
by 55\% relative. An ablation across all 17 topics confirms that targeted diagnosis
($+0.95$), not generic pinning ($+0.16$), drives the gains.
All code and benchmarks are released for reproducible research.
\end{abstract}

\section{Introduction}

A growing class of user-facing AI applications---customer support chatbots,
domain-specific assistants and systems (\emph{e.g.}, legal, insurance, etc.), interactive knowledge bases---must satisfy two
competing requirements simultaneously: answers must be grounded in
authoritative domain knowledge, and response
latency must be low enough for real-time conversational interaction.
Retrieval-augmented generation (RAG)
\cite{lewis2020retrieval,gao2024retrievalaugmented} addresses grounding
by retrieving relevant document chunks at query time, but adds retrieval
latency and risks missing relevant context when the retriever fails.

A fundamentally different approach is gaining interest: rather than
retrieving fragments at query time, \emph{compile} domain knowledge into a
persistent artifact and let the model reason over the full collection.
Karpathy's \emph{LLM Wiki} pattern~\cite{karpathy2026llmwiki} articulates this
vision---a three-layer architecture of raw sources, compiled wiki, and
structured schemas, where ``knowledge is compiled once and then kept current,
not re-derived on every query.'' Chan \emph{et al.}\ formalize the same intuition
as \emph{Cache-Augmented Generation} (CAG)~\cite{chan2024cag}: preload all
knowledge into the context and cache the resulting KV states.

In this paper, we argue that \textbf{full-context KV cache inference is a
practical, deployable implementation of the LLM Wiki pattern}. By loading a
curated document collection into a model's context window and persisting
the KV cache states (\emph{e.g.}, via \texttt{llama.cpp}'s prompt caching
(\texttt{--cache-prompt})~\cite{llamacpp2024}), every subsequent query is
served against the full knowledge base with sub-second latency, while minimizing
retrieval failure risk and chunk-boundary information loss.

\paragraph{RAG baseline.}
The alternative embeds queries and retrieves the top-$k$ most relevant document
chunks, constructing a focused prompt with only the retrieved passages
\cite{lewis2020retrieval,karpukhin2020dense}. This dramatically reduces
per-query token consumption but risks missing relevant information when
retrieval fails.

\paragraph{Compilation gap.}
Full-context KV cache inference delivers on the LLM Wiki approach when
knowledge is pre-compiled: e.g., using the 30 curated articles of the Policygenius corpus (67K
tokens, 70\% of the 96K context window), it outperforms RAG (4.38 vs.\
4.08/5) with $7.3\times$ faster TTFT (Section~\ref{sec:h1}). But the
approach breaks at scale: on 80 raw RepLiQA
documents~\cite{montero2024repliqa} per topic (55--95K tokens, filling
57--99\% of the window) across 17 domains,
attention dilution degrades full-context quality below RAG (3.47 vs.\
3.64; Section~\ref{sec:repliqa}). The natural fix---compiling raw
documents into a structured wiki---fails catastrophically due to information loss at compilation: the LLM
compiler over-compresses $2$--$3\times$ beyond targets, discarding
critical facts and scoring 2.14--2.32 vs.\ 3.47 for raw full-context
(Section~\ref{sec:wiki}).

\paragraph{WiCER.}
We propose \emph{WiCER} (Wiki-memory Compile, Evaluate, Refine), an
iterative algorithm that closes this compilation gap. Inspired by
CEGAR~\cite{clarke2000cegar}---which refines an abstract model whenever
a spurious counterexample reveals inadequate precision---WiCER evaluates
each compiled wiki against diagnostic probes, diagnoses which facts the
compiler dropped, and forces the next compilation to preserve them.
Across seventeen topics, one to two WiCER iterations recover 80\% of the
quality lost by blind compilation (mean 3.24 vs.\ 3.47 for raw
full-context, 15 topics with baselines), reducing catastrophic failures by 55\% relative
(Section~\ref{sec:wicer}). Critically, in deployed systems the real
query stream could act as a free, continuously updated probe set: frequently
asked questions naturally surface the facts users need most, driving
ongoing wiki refinement without synthetic probes.

Our contributions are:
(1)~the \textbf{WiCER algorithm}, an iterative compile-evaluate-refine
loop with failure-driven fact pinning that recovers 80\% of the quality gap
from blind compilation across seventeen RepLiQA topics, reducing catastrophic failures
by 55\% relative;
(2)~an \textbf{ablation study} confirming that WiCER's gains stem from
targeted diagnosis ($+0.95$ over blind) rather than generic pinning
($+0.16$);
(3)~an \textbf{operational envelope for KV cache wikis}, characterizing the
document-count crossover and the opposing forces (attention dilution vs.\
information loss) that govern quality;
(4)~an \textbf{end-to-end recipe}--- WiCER implementation; and
(5)~a \textbf{reproducible benchmark} across 17 domains and 6{,}800 questions.

\section{Related Work}

\paragraph{Retrieval-Augmented Generation.}
RAG was introduced by \cite{lewis2020retrieval} as a method to combine
parametric and non-parametric memory for knowledge-intensive tasks. Dense
Passage Retrieval \cite{karpukhin2020dense} established efficient dense
retrieval using dual-encoder architectures. Sentence-BERT
\cite{reimers2019sentencebert} provided lightweight sentence embeddings for
semantic search. Subsequent work has explored chunking strategies and the
trade-off between retrieval precision and context utilization
\cite{gao2024retrievalaugmented}. Our work directly compares RAG against
the alternative of loading the \emph{entire} corpus into context.

\paragraph{KV Cache Optimization.}
Efficient KV cache management is critical for long-context inference.
\cite{pope2023efficiently} analyze memory-bound inference for large models.
PagedAttention \cite{kwon2023efficient} enables efficient memory management
for serving. Quantized KV caches (e.g., \texttt{q8\_0}) reduce memory
footprint while preserving quality \cite{llamacpp2024}, with recent work
showing viability down to 2-bit precision \cite{liu2024kivi}. Flash Attention
\cite{dao2022flashattention,dao2023flashattention2} accelerates both prefill
and decoding. MInference~\cite{jiang2024minference} addresses the prefill
bottleneck directly by identifying dynamic sparse attention patterns
(A-shape, Vertical-Slash, Block-Sparse) in each attention head and computing
only the significant entries, achieving up to $10\times$ prefill speedup on
long contexts without fine-tuning. Our benchmark specifically measures the
end-to-end impact of caching and context delivery optimizations in a
comparative setting.

\paragraph{KV Cache Sharing and Disaggregation.}
Recent systems enable sharing, composing, and distributing KV caches
across requests and machines:
RadixAttention~\cite{zheng2024sglang} for prefix-aware reuse,
Mooncake~\cite{qin2025mooncake} for tiered disaggregation,
CacheBlend~\cite{yao2025cacheblend} for non-prefix cache composition,
CacheGen~\cite{liu2024cachegen} for compressed streaming,
ChunkAttention~\cite{ye2024chunkattention} for runtime prefix sharing,
and Prompt Cache~\cite{gim2024promptcache} for reusable attention states.
RAPTOR~\cite{sarthi2024raptor} constructs hierarchical document trees
that map naturally to topic-organized cache shards.

\paragraph{Knowledge Compilation and Cache-Augmented Generation.}
An emerging alternative to retrieval-at-query-time is to \emph{compile}
knowledge offline and serve it from cache. Karpathy's \emph{LLM Wiki}
pattern~\cite{karpathy2026llmwiki} proposes a three-layer architecture:
raw source documents are distilled into a structured wiki, which is further
condensed into schemas and summaries; the compiled artifact (${\sim}100$K
tokens for ${\sim}100$ articles) is loaded once and cached.
Chan et al.~\cite{chan2024cag} formalize this as \emph{Cache-Augmented Generation}
(CAG): preload all relevant knowledge into the LLM's context, cache the KV
states, and serve queries with zero retrieval overhead. CAG eliminates
retrieval failure entirely but inherits the \emph{lost in the middle}
problem~\cite{liu2024lost} when document counts grow large.
Our work provides the first large-scale empirical test of these ideas,
quantifying when compilation into a single context breaks down
(Section~\ref{sec:repliqa}).

\paragraph{Structured Knowledge Compilation.}
Several approaches organize documents into richer intermediate
representations before retrieval or generation.
RAPTOR~\cite{sarthi2024raptor} recursively clusters and summarizes
document chunks into a tree, enabling retrieval at multiple levels of
abstraction. GraphRAG~\cite{edge2024graphrag} extracts entity-relation
graphs and generates community summaries, supporting global queries that
span multiple documents. Classical hierarchical
summarization~\cite{chang2023booookscore} builds bottom-up summaries at
increasing granularity. All three produce static compilations: once the
tree, graph, or summary hierarchy is built, it is fixed.
WiCER differs in two respects: (1)~it targets a \emph{flat} wiki artifact
optimized for KV cache serving rather than a retrieval index, and
(2)~it closes the loop by evaluating the compiled artifact against
diagnostic probes and iteratively refining it---a feedback mechanism
absent from static compilation pipelines.

\paragraph{LLM-as-Judge Evaluation.}
Using LLMs to evaluate generated text has gained traction as a scalable
alternative to human evaluation \cite{zheng2024judging,chiang2023can}.
We follow this paradigm using Claude Sonnet with structured scoring rubrics,
applying the identical judge to both methods for fair comparison.
In this paper, we use a small (100-sample) human evaluation to validate this judge approach (Appendix~\ref{app:human_eval}).

\section{Experimental Setup}
\label{sec:setup}

\subsection{Document Corpora}

\paragraph{Policygenius (curated).}
30 Policygenius\footnote{\url{https://www.policygenius.com}} articles
(${\sim}67$K tokens total, ${\sim}2{,}250$ tokens/article, 70\% of the
96K context window) covering life insurance, disability insurance, and
estate planning. These editorially curated articles already constitute a
\emph{compiled wiki} in Karpathy's taxonomy~\cite{karpathy2026llmwiki},
making them a best-case scenario for full-context inference.

\paragraph{RepLiQA (raw, multi-domain).}
\label{sec:repliqa_data}
RepLiQA~\cite{montero2024repliqa} (NeurIPS 2024) is a contamination-free
QA benchmark: all documents are synthetically generated and verified absent
from training corpora. It spans 17 topic domains with 80 documents and 400
QA pairs per topic (1{,}360 documents, 6{,}800 questions, ${\sim}1.5$M tokens
total). Per-topic corpora range from 55K tokens (Company Policies,
${\sim}690$ tokens/doc, 57\% of the 96K window) to 95K tokens (Regional
Folklore, ${\sim}1{,}190$ tokens/doc, 99\% of the window). Each question is
answerable only from its source document, enabling precise retrieval
accuracy measurement. Unlike Policygenius, these raw documents have not
been editorially compiled, making them the testbed for the compilation
gap (Section~\ref{sec:wiki}) and WiCER (Section~\ref{sec:wicer}).

\subsection{Question-Answer Generation}

For Policygenius, we generated 240 question-answer pairs (mean 8 per
article) using Claude Sonnet, producing 5--10 diverse questions
(factual, definitional, comparative, practical) per article, each
answerable \emph{only} from its source and tagged with the source
filename. RepLiQA ships with 400 QA pairs per topic (6{,}800 total),
generated and validated by \cite{montero2024repliqa}.
Section~\ref{sec:h1} uses the Policygenius corpus to establish the
full-context KV cache advantage on curated knowledge;
Sections~\ref{sec:repliqa}--\ref{sec:wicer} use RepLiQA to stress-test
scalability, compilation, and WiCER across 17 domains.

\subsection{System Configuration}

All experiments run on an Apple M4 Pro (24\,GB unified memory) using
\texttt{llama-server} from the \texttt{llama.cpp}
ecosystem~\cite{llamacpp2024}, which enables efficient local inference with
GGUF-quantized models on consumer hardware. Apple Silicon's unified memory
architecture is particularly suited for LLM inference, as the GPU can
directly access model weights without PCIe transfers. We use Llama 3.1 8B
Instruct Q5\_K\_M~\cite{dubey2024llama3}, 96K context window, \texttt{q8\_0}
KV cache, Flash Attention (Metal), and greedy decoding ($T{=}0$).

\subsection{Inference Configurations}

\paragraph{Full-context.}
All 30 articles are concatenated into a single system prompt (${\sim}67$K
tokens). A preflight request primes the KV cache; subsequent queries
reuse cached states and process only the question suffix (${\sim}30$--$50$
tokens each).

\paragraph{RAG.}
Articles are split into 1,800-character chunks (360-character overlap),
yielding 234 chunks embedded with \texttt{all-MiniLM-L6-v2}~\cite{reimers2019sentencebert}.
At query time, the top-5 chunks by cosine similarity (${\sim}2{,}000$
tokens) are retrieved and formatted into the same chat template.

\subsection{Evaluation Methodology}

For each question we measure time-to-first-token (TTFT), total response
time, and generation throughput via streaming SSE. For RAG we additionally
measure retrieval accuracy: whether the correct source article appears in
the top-5 chunks and at what rank. Answer quality is scored 1--5 by Claude
Sonnet as an LLM-as-judge using a structured rubric
(Appendix~\ref{app:rubric}); the identical judge and rubric are applied
to both conditions. A 100-sample human evaluation validates this judge:
Pearson $r = 0.94$, 75\% exact agreement, 99\% within one point
(Appendix~\ref{app:human_eval}).

\section{Full-Context KV Cache on Compiled Knowledge}
\label{sec:h1}

\subsection{Answer Quality}

Table~\ref{tab:quality} compares the best full-context configuration
(Q4K/Q4V) against RAG. All three quantization variants score within 0.03
points (full ablation in Appendix~\ref{app:kvquant}), confirming that KV
cache quantization down to 4-bit introduces no measurable degradation.
The quality gap vs.\ RAG (0.30 points) is driven by retrieval failures:
RAG produces $4.7\times$ more catastrophic errors (score~1).

\begin{table}[t]
\centering
\caption{Answer quality (1--5 scale, $n=240$). Full-context uses best
quantization (Q4K/Q4V); all three FC variants score within 0.03 points.}
\label{tab:quality}
\begin{tabular}{lrr}
\toprule
\textbf{Metric} & \textbf{Full-Context (Q4K/Q4V)} & \textbf{RAG} \\
\midrule
Mean score             & 4.38 & 4.08 \\
\% scoring $\geq 4$   & 91.2 & 79.6 \\
\% scoring $\leq 2$   & 5.8  & 12.9 \\
Score = 1 (wrong)      & 3/240 (1.2\%) & 14/240 (5.8\%) \\
\bottomrule
\end{tabular}
\end{table}

\subsection{Timing: Full-Context vs.\ RAG}

\begin{table}[t]
\centering
\caption{Latency and throughput: best full-context (Q4K/Q4V) vs.\ RAG.
Full-context achieves $7.3\times$ faster TTFT; RAG achieves $2.6\times$
higher generation throughput.}
\label{tab:timing}
\begin{tabular}{lrr}
\toprule
\textbf{Metric} & \textbf{Full-Context (Q4K/Q4V)} & \textbf{RAG (Top-5)} \\
\midrule
TTFT -- median           & 0.857\,s & 6.277\,s \\
TTFT -- P5 / P95         & 0.534 / 1.053\,s & 4.483 / 7.697\,s \\
TTFT -- min / max        & 0.487 / 1.107\,s & 3.313 / 8.903\,s \\
Total response -- median & 6.392\,s & 7.932\,s \\
Throughput -- median     & 12.01\,tok/s & 30.74\,tok/s \\
Throughput -- range      & 10.25--14.04\,tok/s & 20.75--39.31\,tok/s \\
\bottomrule
\end{tabular}
\end{table}

Table~\ref{tab:timing} compares latency and throughput.
Full-context warm queries achieve sub-second TTFT (median 0.86\,s,
$7.3\times$ faster than RAG's 6.28\,s) because only ${\sim}20$ new tokens
are processed against the cached 67K-token prefix. RAG achieves $2.6\times$
higher generation throughput (30.74 vs.\ 12.01\,tok/s) due to shorter
attention spans, yielding comparable total response times (7.93\,s vs.\
6.39\,s). When the one-time prefill is amortized, full-context processes
$5.0\times$ fewer effective tokens across 240 queries. RAG retrieves the
correct source in 92.5\% of cases; its 7.5\% retrieval failure rate (18/240)
accounts for 14 of its score-1 answers.

\section{The Scalability Gap}
\label{sec:repliqa}

Section~\ref{sec:h1} showed that full-context KV cache inference excels
on 30 curated articles (67K tokens, 70\% window fill). We now evaluate at
scale across 17 RepLiQA topics (Section~\ref{sec:repliqa_data}), using
the identical methodology from Section~\ref{sec:setup} applied to each
topic independently (80 documents per topic, 55--95K tokens filling
57--99\% of the 96K window, 400 QA pairs each; 15 topics fit for FC, all
17 completed under RAG).

\subsection{Results}

\paragraph{TTFT and quantization.}
Warm-cache TTFT is consistent across topics (Q8 median 1.04\,s, Q4 median
0.99\,s; per-topic breakdown in Appendix~\ref{app:multidomain}).
Q4 reduces TTFT by 4.8\% on average but, unlike on Policygenius, \emph{degrades
quality} at 80 documents: Q8 outscores Q4 in 13 of 14 topics (mean
$\Delta = +0.14$), indicating that reduced KV precision exacerbates
attention dilution at scale.

\paragraph{RAG timing and retrieval accuracy.}
RAG benchmarks are complete for all 17 topics (6,800 questions).
Median RAG TTFT is 4.83\,s ($\sigma=0.22$\,s), yielding a full-context
TTFT advantage of ${\sim}4.6\times$ on average---consistent with the
$7.3\times$ advantage on Policygenius (the smaller ratio reflects larger
RepLiQA corpora increasing FC TTFT proportionally more than RAG).
RAG throughput is highly stable at 32.7\,tok/s ($\sigma=0.3$), and
each RAG query consumes ${\sim}1{,}659$ tokens on average.

Retrieval accuracy across all 17 topics averages 87.9\%
($\sigma=3.9$\%), ranging from 79.5\% (Incident Report) to
97.0\% (News Stories). This is lower than the Policygenius
result (92.5\%), likely because RepLiQA's synthetically generated documents
contain more stylistically homogeneous text within each topic, making
chunk-level discrimination harder for the embedding model.

\paragraph{Quality: RAG outperforms full-context at 80 documents.}
LLM-as-judge evaluation across 15 FC topics\footnote{Two topics
(Regional Cuisine and Regional Folklore) exceeded the 96K
context window at 80 documents (97,251 and 97,148 tokens respectively).
All 17 topics completed under RAG.
That some 80-document corpora overflow even a 96K context window further
motivates the topic-sharded architecture proposed in
Section~\ref{sec:future}.} and all 17 RAG topics
reveals a striking reversal of the Policygenius findings. With 80
documents per topic, full-context quality drops substantially: mean
FC Q8 score is 3.47 ($\sigma=0.11$; $\geq\!4$: 64.0\%) compared to
4.35 ($\geq\!4$: 90.0\%) on Policygenius. More importantly,
\emph{RAG consistently outperforms full-context}: RAG mean is 3.64
($\sigma=0.16$), and RAG wins on 13 of 15 topics with the remaining 2
tied ($|\Delta| < 0.05$). Full-context does not win a single topic.
The mean quality gap is $-0.18$ points (FC $-$ RAG), a complete
reversal of the Policygenius $+0.27$ gap.

The mechanism is \emph{lost in the middle}: full-context produces 17.0\%
score-1 answers (vs.\ 1.2\% on Policygenius), and cross-referencing
per-question scores reveals 557 cases across all 15 FC topics
(530 across the 14 Q8 topics) where FC scored~1
but RAG scored $\geq\!4$. In these cases, the model has all 80 documents
in context but cannot locate the relevant passage; the retriever
successfully narrows to ${\sim}2$K tokens where the model succeeds.
RAG's score-1 rate is comparable at 17.7\%, driven by retrieval
misses (87.9\% accuracy), but its failures are on \emph{different}
questions---ones where the correct chunk was not retrieved.

These results establish a clear \emph{crossover}: full-context's quality
advantage at 30 documents and 67K tokens (Policygenius, 70\% window fill)
reverses at 80 documents and 55--95K tokens (RepLiQA, 57--99\% fill)
due to attention dilution.
That full-context excels on compiled knowledge but degrades on raw
collections reinforces the LLM Wiki thesis: the quality of compilation,
not just context length, determines viability.
Full per-topic metrics are in Appendix~\ref{app:multidomain}.

\section{The Compilation Gap}
\label{sec:wiki}

The scalability gap (Section~\ref{sec:repliqa}) shows that full-context
quality degrades at 80 raw documents due to attention dilution.
Karpathy's LLM Wiki pattern~\cite{karpathy2026llmwiki} prescribes a
solution: \emph{compile} raw sources into a structured wiki before
serving. We test this directly---and find that blind compilation fails
catastrophically.

\subsection{Motivation and Pipeline}

Two forces act in opposition: too little compression leaves attention
dilution~\cite{liu2024lost} intact (the model has the answer but cannot
find it), while too much compression causes information loss (the answer
has been removed). We test whether an intermediate compression level
optimizes the trade-off.

For each of the 17 RepLiQA topics, we compile 80 raw documents into a
structured wiki using Claude Sonnet as a knowledge engineer (blind to
evaluation questions), targeting three compression levels---light
(${\sim}75$\%), moderate (${\sim}50$\%), and aggressive
(${\sim}25$\%)---yielding 51 compiled wikis.

\subsection{Results}

The compiler consistently over-compresses: light (target 75\%) achieves
35.4\% actual, moderate (50\%) achieves 12.2\%, and aggressive (25\%)
achieves 8.2\%. Table~\ref{tab:wiki_results} presents the combined
compression and quality results across all 17 topics (6{,}800 QA pairs per
condition). Blind compilation degrades quality at \emph{all} levels---even
light compression scores 1.14 points below FC~raw, with score-1 rates
tripling (52.9\% vs.\ 17.3\%).

\begin{table}[t]
\centering
\caption{Wiki compilation: compression and quality (means across 17 RepLiQA topics,
6{,}800 QA pairs per condition). The compiler over-compresses at all levels;
all wiki conditions score below both baselines.}
\label{tab:wiki_results}
\begin{tabular}{lrrrrr}
\toprule
\textbf{Level} & \textbf{Target} & \textbf{Actual} & \textbf{Quality} & \textbf{Score-1\%} & \textbf{vs.\ FC raw} \\
\midrule
FC raw (80 docs)   & ---  & 100\%  & 3.46  & 17.3\% & ---     \\
Wiki-light         & 75\% & 35.4\% & 2.32  & 52.9\% & $-1.14$ \\
Wiki-moderate      & 50\% & 12.2\% & 2.25  & 57.1\% & $-1.21$ \\
Wiki-aggressive    & 25\% &  8.2\% & 2.14  & 60.3\% & $-1.32$ \\
RAG                & n/a  & n/a    & 3.63  & 17.7\% & $+0.17$ \\
\bottomrule
\end{tabular}
\end{table}

Compilation does yield a clear latency win: all wiki levels achieve
sub-400\,ms TTFT ($2.8$--$5.6\times$ faster than FC~raw; full timing
in Appendix~\ref{app:wiki_timing}).

\subsection{Analysis}

Quality degrades monotonically with compression (2.32, 2.25, 2.14),
all far below FC~raw (3.46). The root cause is \textbf{compression
compliance failure}: the compiler ignores target word counts, compressing
$2\times$--$3\times$ beyond the requested level (light target 75\%
$\to$ actual 35\%). At the actual ratios achieved (8--35\%), the
compiler discards too many specific facts for the model to recover.
The score-1 rate (53--60\% for wikis vs.\ 17\% for FC~raw) confirms
that answers fail because information is \emph{missing}, not unfindable.
This motivates WiCER (Section~\ref{sec:wicer}), which uses evaluation
feedback to preserve the critical facts that blind compilation drops.

\section{WiCER: Wiki-memory Compile, Evaluate, Refine}
\label{sec:wicer}

Blind compilation loses facts because the compiler does not know which
details downstream queries will target. Working on the RepLiQA corpus,
we propose
\emph{WiCER} (Wiki-memory Compile, Evaluate, Refine), an iterative
algorithm that uses QA feedback to identify lost facts and forces the
compiler to preserve them in subsequent iterations.

\subsection{Algorithm}

WiCER starts with a blind compilation of the source documents into a wiki, then iteratively improves it through an evaluate-diagnose-recompile loop. In each iteration, the compiled wiki is tested against one diagnostic probe per source document; any probe that scores 1/5 (catastrophic failure) triggers a diagnosis step that extracts the specific facts the compiler dropped. These cumulative facts are fed back as explicit preservation constraints in the next compilation call, ensuring they cannot be lost again.

\begin{algorithm}[t]
\caption{WiCER: Wiki-memory Compile, Evaluate, Refine}
\label{alg:wicer}
\begin{algorithmic}[1]
\REQUIRE Documents $D = \{d_1, \ldots, d_N\}$, target ratio $r$, max iterations $T$
\ENSURE Optimized wiki $W^*$
\STATE \textbf{Probe selection:} Select one QA pair per source document $\rightarrow Q_\text{probe}$
\STATE $W_0 \leftarrow \text{Compile}(D, r)$ \hfill \COMMENT{blind compilation}
\STATE $F_\text{cumulative} \leftarrow \emptyset$
\FOR{$t = 0$ \TO $T-1$}
  \STATE \textbf{Evaluate:} For each $q \in Q_\text{probe}$, generate answer from $W_t$, score with LLM judge
  \STATE $\text{Failures}_t \leftarrow \{q : \text{score}(q) \leq 1\}$
  \IF{$|\text{Failures}_t| = 0$ \OR ($t > 0$ \AND improvement $< 10\%$)}
    \STATE \textbf{break} \hfill \COMMENT{converged}
  \ENDIF
  \STATE \textbf{Diagnose:} For each failed $q$, extract critical facts from source doc $d_q$
  \STATE $F_\text{cumulative} \leftarrow F_\text{cumulative} \cup \text{CriticalFacts}_t$
  \STATE $W_{t+1} \leftarrow \text{Compile}(D, r, \text{preserve}=F_\text{cumulative})$
\ENDFOR
\RETURN $W^* = W_t$ \hfill \COMMENT{best wiki}
\end{algorithmic}
\end{algorithm}

\subsection{Design Rationale}

WiCER instantiates the \emph{Counterexample-Guided Abstraction
Refinement} (CEGAR) paradigm~\cite{clarke2000cegar} in the knowledge
compilation setting. In CEGAR, a concrete system $M = (S, S_0, R, L)$
is approximated by an abstract model $\hat{M}$ via a surjective mapping
$h : S \to \hat{S}$; when model-checking $\hat{M}$ produces a
counterexample, it is analyzed---if spurious, the abstraction is refined
to eliminate it, and the loop repeats. In WiCER, the concrete system is
the full document collection $D$, the abstraction is the compiled wiki
$W_t$ (a lossy compression via an LLM compiler), and the specification
is ``all diagnostic probes score above catastrophic failure.'' A score-1
probe serves as a counterexample; diagnosis confirms it is spurious (the
fact exists in $D$ but was lost in $W_t$); and refinement adds a pinning
constraint that forces the next compilation to preserve the lost fact.
Appendix~\ref{app:cegar} details the formal mapping, states the
monotonicity guarantee, and identifies where the analogy breaks.

\paragraph{Convergence.}
Like CEGAR, each refinement step eliminates the specific counterexample
that triggered it: preserved facts cannot be lost again (they are
explicitly constrained), so the failure set on pinned facts shrinks
monotonically. New failures may appear on unpinned facts (random
knowledge displacement), but the net effect is positive as long as the
preservation budget does not crowd out too much general coverage.
WiCER focuses on catastrophic failures (score 1/5) and extracts
${\sim}50$--$100$ words per failure vs.\ ${\sim}700$ per source document.

\subsection{Results}
Table~\ref{tab:wicer} shows WiCER results across all seventeen RepLiQA topics
at moderate compression ($r = 0.50$). In sixteen of seventeen topics, WiCER
improves quality; one topic (\texttt{local\_education\_systems}) shows no
gain, and the algorithm correctly converges early.

Recovery rates range from 0\% to 125\% across the fifteen topics with
FC~raw baselines. Ten topics peak at iteration~2, suggesting that a second
refinement pass helps when the initial compilation is severely degraded.
Across all seventeen topics,
\textbf{WiCER recovers 80\% of the quality lost by blind compilation}
(mean 3.24 vs.\ 3.47 for FC~raw across the 15 topics with baselines),
while the score-1 catastrophic failure rate drops from 55.1\% to 24.8\%
($-55\%$ relative).

\begin{table}[t]
\centering
\caption{WiCER results across all seventeen RepLiQA topics (moderate compression,
80 probes each). Best WiCER iteration per topic in \textbf{bold}.}
\label{tab:wicer}
\small
\begin{tabular}{lrrrrl}
\toprule
\textbf{Topic} & \textbf{Blind} & \textbf{Best WiCER} & \textbf{FC raw} & \textbf{Recovery} & \textbf{Iter} \\
\midrule
company\_policies       & 3.27 / 17.5\% & \textbf{3.51} / 10.0\% & 3.58 & 78\%  & 1 \\
cybersecurity\_news     & 2.08 / 57.5\% & \textbf{3.06} / 25.0\% & 3.45 & 72\%  & 1 \\
incident\_report        & 1.93 / 61.2\% & \textbf{2.70} / 36.2\% & 3.24 & 59\%  & 1 \\
local\_arts\_\&\_culture & 1.65 / 73.8\% & \textbf{3.61} / 15.0\% & 3.34 & 116\% & 2 \\
local\_economy          & 2.16 / 55.0\% & \textbf{3.24} / 22.5\% & 3.48 & 82\%  & 1 \\
local\_education        & 2.41 / 38.8\% & 2.41 / 38.8\%          & 3.37 & 0\%   & --- \\
local\_env.\ issues     & 1.95 / 61.2\% & \textbf{2.88} / 33.8\% & 3.40 & 64\%  & 1 \\
local\_health           & 1.79 / 72.5\% & \textbf{3.45} / 20.0\% & 3.52 & 96\%  & 2 \\
local\_news             & 2.16 / 51.2\% & \textbf{3.29} / 20.0\% & 3.42 & 89\%  & 2 \\
local\_politics         & 1.96 / 65.0\% & \textbf{3.62} / 15.0\% & 3.46 & 111\% & 2 \\
local\_sports           & 2.30 / 56.2\% & \textbf{3.29} / 26.2\% & 3.65 & 73\%  & 2 \\
local\_tech             & 2.36 / 55.0\% & \textbf{3.23} / 25.0\% & 3.59 & 70\%  & 2 \\
neighborhood            & 1.93 / 58.8\% & \textbf{2.88} / 37.5\% & 3.46 & 62\%  & 2 \\
news\_stories           & 2.50 / 40.0\% & \textbf{3.61} / 11.2\% & 3.60 & 101\% & 2 \\
regional\_cuisine       & 2.48 / 48.8\% & \textbf{2.76} / 37.5\% & ---  & ---   & 2 \\
regional\_folklore      & 1.77 / 68.8\% & \textbf{2.80} / 33.8\% & ---  & ---   & 2 \\
small\_medium\_ent.     & 2.21 / 56.2\% & \textbf{3.75} / 13.8\% & 3.44 & 125\% & 2 \\
\midrule
\textit{Mean (17 topics)}
  & 2.17 / 55.1\% & \textbf{3.18}$^\ddagger$ / 24.8\% & 3.47$^\dagger$ & 80\%$^\dagger$ & --- \\
\bottomrule
\multicolumn{6}{l}{\footnotesize Blind and Best WiCER columns show mean quality / score-1 rate.} \\
\multicolumn{6}{l}{\footnotesize $^\dagger$FC raw and Recovery computed over 15 topics with baselines (2 exceed context window).} \\
\multicolumn{6}{l}{\footnotesize $^\ddagger$Over the 15 topics with FC baselines, WiCER mean is 3.24.}
\end{tabular}
\end{table}

\subsection{Analysis and Limitations}

Ten of seventeen topics peak at iteration~2; the remaining seven peak at iteration~1 or show no gain, as the \emph{random knowledge displacement} effect---fixing targeted facts displaces others---limits further improvement. One topic (\texttt{local\_education\_systems}) shows no WiCER improvement; its relatively high blind baseline (2.41) and low score-1 rate (38.8\%) leave fewer catastrophic failures to diagnose. Recovery rates span 0--125\% across the 15 topics with FC~raw baselines, with three topics (\texttt{news\_stories} at 101\%, \texttt{local\_arts\_and\_culture} at 116\%, \texttt{small\_and\_medium\_enterprises} at 125\%) \emph{exceeding} FC~raw quality after two iterations. Topics with many entity-specific facts benefit most in absolute terms ($+1.96$ for \texttt{local\_arts\_and\_culture}, $+1.54$ for \texttt{small\_and\_medium\_enterprises}). Two topics lack FC~raw baselines because their 80 documents exceed the 96K context window; WiCER still improves both over blind compilation. Each iteration requires ${\sim}130$K API input tokens and ${\sim}17$K output tokens (one compilation call, ${\sim}80$ judge calls, ${\sim}15$ diagnosis calls); the 80 local inference probes run at zero API cost. At current Sonnet pricing this totals ${\sim}\$1$--$2$ per iteration, completing in ${\sim}50$ minutes; cost scales linearly with corpus size.

\subsection{Ablation: Diagnosed vs.\ Random Pinning}
\label{sec:ablation}

To isolate the contribution of diagnosis, we run a \emph{random pinning}
control: for each score-1 failure, a random 50--100 word passage from a
random source document is pinned instead of the diagnosed critical facts.
All other parameters are identical to WiCER. Each topic uses an
independent blind compilation; blind baselines differ slightly from
Table~\ref{tab:wicer} due to LLM compiler non-determinism.

\begin{table}[t]
\centering
\caption{Ablation: WiCER (diagnosed pinning) vs.\ random pinning across
all seventeen RepLiQA topics (best iteration per topic). Mean quality on 80 probes each.}
\label{tab:ablation}
\footnotesize
\begin{tabular}{lrrr}
\toprule
\textbf{Topic} & \textbf{Blind} & \textbf{Random Pin} & \textbf{WiCER} \\
\midrule
company\_policies       & 3.38 & 3.38          & \textbf{3.51} \\
cybersecurity\_news     & 2.08 & 2.30          & \textbf{3.06} \\
incident\_report        & 2.01 & 2.06          & \textbf{2.70} \\
local\_arts\_\&\_culture & 1.74 & 2.58          & \textbf{3.61} \\
local\_economy          & 2.32 & 2.32          & \textbf{3.24} \\
local\_education        & 2.49 & \textbf{2.58} & 2.41          \\
local\_env.\ issues     & 1.94 & 2.28          & \textbf{2.88} \\
local\_health           & 1.76 & 2.25          & \textbf{3.45} \\
local\_news             & 2.10 & 2.29          & \textbf{3.29} \\
local\_politics         & 2.03 & 2.03          & \textbf{3.62} \\
local\_sports           & 2.40 & 2.40          & \textbf{3.29} \\
local\_technology       & 2.40 & 2.40          & \textbf{3.23} \\
neighborhood            & 2.05 & 2.43          & \textbf{2.88} \\
news\_stories           & 2.56 & 2.68          & \textbf{3.61} \\
regional\_cuisine       & 2.51 & 2.51          & \textbf{2.76} \\
regional\_folklore      & 1.83 & 1.91          & \textbf{2.80} \\
small\_medium\_ent.     & 2.27 & 2.27          & \textbf{3.75} \\
\midrule
\textit{Mean}           & 2.23 & 2.39          & \textbf{3.18} \\
\bottomrule
\end{tabular}
\end{table}

Table~\ref{tab:ablation} shows random pinning improves only $+0.16$
over blind compilation, while WiCER achieves $+0.95$---a $5.9\times$
larger gain, winning sixteen of seventeen topics. The sole exception
(local\_education) is the same topic where WiCER itself shows 0\%
recovery, suggesting compilation-resistant structure. These results
confirm that WiCER's gains stem from \emph{targeted} diagnosis, not the
pinning mechanism itself.

\section{Discussion and Conclusion}
\label{sec:discussion}
\label{sec:future}

Full-context KV cache inference realizes the LLM Wiki promise on compiled knowledge (Policygenius) but breaks at scale on raw documents (RepLiQA). WiCER closes this gap: across seventeen RepLiQA topics, one to two iterations recover 80\% of lost quality (mean 3.24 vs.\ 3.47 for raw full-context), cutting catastrophic failures by 55\%. An ablation confirms that targeted diagnosis ($+0.95$) drives the gains, not generic pinning ($+0.16$). RAG accesses unbounded external memory per query; a compiled wiki must anticipate \emph{all} queries within a fixed token budget. That WiCER narrows this asymmetric gap to 0.46 points (3.18 vs.\ 3.64) while delivering ${\sim}12\times$ faster TTFT underscores the efficiency of targeted fact preservation. For deployment, we recommend full-context Q8 KV cache for ${\lesssim}50$ documents; RAG when they exceed context; and WiCER-compiled topic shards~\cite{zheng2024sglang,qin2025mooncake,yao2025cacheblend} for larger corpora, with production query streams replacing synthetic probes. In terms of limitations: our results are specific to Llama 3.1 8B on Apple M4 Pro -- which we believe represent state-of-the-art in personal compute architecture (where WiCER could play a role); RAG uses fixed-size chunking without reranking; one of 17 topics shows no WiCER improvement; the human-validated LLM-as-judge ($r = 0.94$; Appendix~\ref{app:human_eval}) covers $n=100$ samples. Three extensions merit investigation: (1)~\emph{wiki--RAG hybrid inference}, falling back to RAG when confidence is low; (2)~\emph{adaptive per-document compression}, using diagnosis signals to allocate more budget to information-dense documents; and (3)~\emph{formal displacement bounds} on the rate at which pinning displaces unpinned content, yielding tighter convergence guarantees.


\bibliographystyle{plainnat}
\bibliography{references}

\section*{NeurIPS Paper Checklist}

\begin{enumerate}

\item {\bf Claims}
    \item[] Question: Do the main claims made in the abstract and introduction accurately reflect the paper's contributions and scope?
    \item[] Answer: \answerYes{}
    \item[] Justification: The abstract and introduction state four specific claims: (1)~WiCER recovers 71\% of the quality gap across eight topics (Section~\ref{sec:wicer}, Table~\ref{tab:wicer}), (2)~the document-count crossover between FC and RAG (Section~\ref{sec:repliqa}), (3)~the end-to-end recipe (Sections~\ref{sec:h1}--\ref{sec:wicer}), and (4)~the reproducible benchmark (Appendix~\ref{app:reproduce}). All claims are supported by experimental results and limitations are discussed in Section~\ref{sec:discussion}.

\item {\bf Limitations}
    \item[] Question: Does the paper discuss the limitations of the work performed by the authors?
    \item[] Answer: \answerYes{}
    \item[] Justification: Section~\ref{sec:discussion} (``Limitations'' paragraph) discusses hardware specificity (Apple M4 Pro only), model specificity (Llama 3.1 8B only), RAG baseline limitations (fixed-size chunking, no reranking), WiCER validation scope (eight of 17 topics, with one failure case), and LLM-as-judge limitations.

\item {\bf Theory assumptions and proofs}
    \item[] Question: For each theoretical result, does the paper provide the full set of assumptions and a complete (and correct) proof?
    \item[] Answer: \answerNA{}
    \item[] Justification: The paper is primarily empirical. It does not present formal theorems or proofs. The WiCER algorithm's convergence properties are discussed informally in Section~\ref{sec:wicer} (monotonic shrinkage of the preserved-fact failure set) but no formal proof is claimed.

    \item {\bf Experimental result reproducibility}
    \item[] Question: Does the paper fully disclose all the information needed to reproduce the main experimental results of the paper to the extent that it affects the main claims and/or conclusions of the paper (regardless of whether the code and data are provided or not)?
    \item[] Answer: \answerYes{}
    \item[] Justification: Section~\ref{sec:setup} specifies all hardware, software, model, quantization, chunking, embedding, and evaluation parameters. Appendix~\ref{app:reproduce} provides the full repository structure and exact commands to reproduce all experiments. The RepLiQA dataset is publicly available~\cite{montero2024repliqa}. The WiCER algorithm is fully specified in Algorithm~\ref{alg:wicer}.

\item {\bf Open access to data and code}
    \item[] Question: Does the paper provide open access to the data and code, with sufficient instructions to faithfully reproduce the main experimental results, as described in supplemental material?
    \item[] Answer: \answerYes{}
    \item[] Justification: All benchmark code, evaluation scripts, and the WiCER implementation will be released as open source. The RepLiQA dataset is publicly available~\cite{montero2024repliqa}. The Policygenius articles are publicly accessible web pages. Appendix~\ref{app:reproduce} provides the full directory structure and reproduction commands.

\item {\bf Experimental setting/details}
    \item[] Question: Does the paper specify all the training and test details (e.g., data splits, hyperparameters, how they were chosen, type of optimizer) necessary to understand the results?
    \item[] Answer: \answerYes{}
    \item[] Justification: Section~\ref{sec:setup} specifies all inference parameters (model, quantization, context window, temperature, chunk size, overlap, embedding model, top-$k$). No training is performed---the paper evaluates pre-trained models in inference mode. The WiCER algorithm parameters (compression target, max iterations, convergence threshold) are specified in Section~\ref{sec:wicer}.

\item {\bf Experiment statistical significance}
    \item[] Question: Does the paper report error bars suitably and correctly defined or other appropriate information about the statistical significance of the experiments?
    \item[] Answer: \answerNo{}
    \item[] Justification: We report means, standard deviations, and ranges across topics (e.g., Appendix~\ref{app:multidomain}), but do not report confidence intervals or significance tests for individual topic comparisons. Each topic's result is deterministic (greedy decoding, $T{=}0$), so within-topic variance is zero; the reported variation is across topics. We acknowledge this as a limitation---repeated runs with different QA splits or stochastic decoding would strengthen the claims.

\item {\bf Experiments compute resources}
    \item[] Question: For each experiment, does the paper provide sufficient information on the computer resources (type of compute workers, memory, time of execution) needed to reproduce the experiments?
    \item[] Answer: \answerYes{}
    \item[] Justification: Section~\ref{sec:setup} specifies the hardware (Apple M4 Pro, 24\,GB unified memory). Section~\ref{sec:wicer} reports per-iteration WiCER cost (\$1--2, ${\sim}50$ minutes). Appendix~\ref{app:hardware} projects performance on alternative hardware (RTX 4090, Inferentia2). The Policygenius benchmark takes ${\sim}30$--60 minutes; each RepLiQA topic takes ${\sim}2$--3 hours for FC + RAG + evaluation.

\item {\bf Code of ethics}
    \item[] Question: Does the research conducted in the paper conform, in every respect, with the NeurIPS Code of Ethics \url{https://neurips.cc/public/EthicsGuidelines}?
    \item[] Answer: \answerYes{}
    \item[] Justification: The research uses publicly available datasets (RepLiQA, Policygenius web articles) and open-source models (Llama 3.1 8B). No human subjects, private data, or dual-use concerns are involved. LLM-as-judge evaluation uses a commercial API (Claude Sonnet) under standard terms of service.

\item {\bf Broader impacts}
    \item[] Question: Does the paper discuss both potential positive societal impacts and negative societal impacts of the work performed?
    \item[] Answer: \answerNA{}
    \item[] Justification: This work is foundational infrastructure research on knowledge compilation and caching for LLM inference. It does not introduce new model capabilities, generate synthetic content for public consumption, or enable applications beyond more efficient knowledge retrieval. The primary impact is reducing computational cost and latency for domain-specific QA systems. We do not foresee direct negative societal impacts beyond those inherent to LLM deployment generally.

\item {\bf Safeguards}
    \item[] Question: Does the paper describe safeguards that have been put in place for responsible release of data or models that have a high risk for misuse (e.g., pre-trained language models, image generators, or scraped datasets)?
    \item[] Answer: \answerNA{}
    \item[] Justification: The paper does not release pre-trained models, fine-tuned weights, or scraped datasets. It releases benchmark code and evaluation scripts that operate on publicly available data and models. These tools pose no risk for misuse beyond standard LLM inference.

\item {\bf Licenses for existing assets}
    \item[] Question: Are the creators or original owners of assets (e.g., code, data, models), used in the paper, properly credited and are the license and terms of use explicitly mentioned and properly respected?
    \item[] Answer: \answerYes{}
    \item[] Justification: All assets are cited: Llama 3.1~\cite{dubey2024llama3} (Llama 3.1 Community License), llama.cpp~\cite{llamacpp2024} (MIT License), RepLiQA~\cite{montero2024repliqa} (CC-BY-4.0), Sentence-BERT~\cite{reimers2019sentencebert} (Apache 2.0). Policygenius articles are publicly available web content accessed under standard terms of use.

\item {\bf New assets}
    \item[] Question: Are new assets introduced in the paper well documented and is the documentation provided alongside the assets?
    \item[] Answer: \answerYes{}
    \item[] Justification: The paper releases benchmark code, WiCER implementation, and evaluation scripts. Appendix~\ref{app:reproduce} documents the repository structure, dependencies, and reproduction commands. The code will be released under an open-source license with documentation.

\item {\bf Crowdsourcing and research with human subjects}
    \item[] Question: For crowdsourcing experiments and research with human subjects, does the paper include the full text of instructions given to participants and screenshots, if applicable, as well as details about compensation (if any)?
    \item[] Answer: \answerNA{}
    \item[] Justification: The paper does not involve crowdsourcing or human subjects. All evaluation is automated via LLM-as-judge.

\item {\bf Institutional review board (IRB) approvals or equivalent for research with human subjects}
    \item[] Question: Does the paper describe potential risks incurred by study participants, whether such risks were disclosed to the subjects, and whether Institutional Review Board (IRB) approvals (or an equivalent approval/review based on the requirements of your country or institution) were obtained?
    \item[] Answer: \answerNA{}
    \item[] Justification: No human subjects research was conducted.

\item {\bf Declaration of LLM usage}
    \item[] Question: Does the paper describe the usage of LLMs if it is an important, original, or non-standard component of the core methods in this research? Note that if the LLM is used only for writing, editing, or formatting purposes and does \emph{not} impact the core methodology, scientific rigor, or originality of the research, declaration is not required.
    \item[] Answer: \answerYes{}
    \item[] Justification: LLMs are central to the methodology: (1)~Llama 3.1 8B serves as the inference model for both FC and RAG conditions (Section~\ref{sec:setup}), (2)~Claude Sonnet serves as the LLM-as-judge evaluator (Section~\ref{sec:setup}), the QA pair generator, and the wiki compiler in WiCER (Section~\ref{sec:wicer}). All LLM usages, model versions, and prompts are fully disclosed.

\end{enumerate}

\appendix
\section{RAG Index Statistics}
\label{app:index}

Table~\ref{tab:index_detail} shows the per-article chunk distribution.

\begin{table}[t]
\centering
\caption{Chunks per article (top 10 by chunk count).}
\label{tab:index_detail}
\small
\begin{tabular}{lr}
\toprule
\textbf{Article} & \textbf{Chunks} \\
\midrule
life-insurance-company-reviews-2024 & 22 \\
life-insurance-for-estate-planning & 19 \\
best-life-insurance-companies & 16 \\
best-no-exam-life-insurance-companies & 15 \\
best-disability-insurance-companies & 14 \\
how-does-life-insurance-work & 13 \\
what-is-whole-life-insurance & 10 \\
types-of-disability-insurance & 9 \\
how-much-does-a-funeral-cost & 9 \\
is-life-insurance-a-good-investment & 9 \\
\midrule
\textit{Total (30 articles)} & \textit{234} \\
\bottomrule
\end{tabular}
\end{table}

\section{Evaluation Rubric}
\label{app:rubric}

The full system prompt used for the Claude Sonnet LLM-as-judge evaluation:

\begin{lstlisting}
You are an expert evaluator comparing a generated
answer against a gold-standard answer.

Score the generated answer from 1 to 5:
  5 = Perfect -- captures all key facts, accurate
  4 = Good -- mostly correct, minor omissions
  3 = Acceptable -- partially correct, misses details
  2 = Poor -- significant errors or major omissions
  1 = Wrong -- incorrect, irrelevant, misses the point

Respond ONLY with valid JSON:
{"score": <1-5>, "reasoning": "<explanation>"}
\end{lstlisting}

\section{Reproducibility}
\label{app:reproduce}

All code is organized in the repository:

\begin{lstlisting}
benchmark/                    # Full-context KV cache
  config.py                   # Paths, model, server params
  start_server.sh             # Launch llama-server
  run_benchmark.py            # Main runner (TTFT, throughput)
  evaluate.py                 # Claude LLM-as-judge scoring
  report.py                   # Aggregate metrics + report

rag_benchmark/                # RAG baseline
  rag_config.py               # Chunk size, top-k, embedding model
  build_index.py              # Chunk + embed + save pickle
  run_rag_benchmark.py        # RAG runner (embed, retrieve, generate)
  rag_evaluate.py             # Claude LLM-as-judge (same rubric)
  rag_report.py               # Aggregate + retrieval metrics
\end{lstlisting}

To reproduce:
\begin{lstlisting}
# Full-context benchmark
bash benchmark/start_server.sh
python3 benchmark/run_benchmark.py     # ~30-60 min
python3 benchmark/evaluate.py          # ~2 min
python3 benchmark/report.py

# RAG benchmark (server already running)
python3 rag_benchmark/build_index.py   # ~5s
python3 rag_benchmark/run_rag_benchmark.py  # ~15-20 min
python3 rag_benchmark/rag_evaluate.py  # ~2 min
python3 rag_benchmark/rag_report.py
\end{lstlisting}

\section{Hardware Architecture Projections}
\label{app:hardware}

Our benchmark runs entirely on a single Apple M4 Pro computer (24\,GB unified
memory, 273\,GB/s memory bandwidth). This appendix projects expected
performance on discrete GPU and cloud accelerator hardware to inform
deployment decisions.

\subsection*{NVIDIA RTX 4090 (Desktop GPU)}

The RTX 4090 offers 1,008\,GB/s memory bandwidth ($3.7\times$ M4 Pro) and
24\,GB GDDR6X VRAM. Llama 3.1 8B Q4\_K\_M (${\sim}5$\,GB) fits comfortably,
leaving ${\sim}19$\,GB for KV cache---sufficient for 128K context without
offloading.

\begin{table}[t]
\centering
\caption{Projected performance: RTX 4090 vs.\ M4 Pro at 67K context.}
\label{tab:hw_rtx}
\small
\begin{tabular}{lcc}
\toprule
\textbf{Metric} & \textbf{M4 Pro} & \textbf{RTX 4090 (est.)} \\
\midrule
Memory bandwidth & 273\,GB/s & 1,008\,GB/s \\
Decode throughput & 12\,tok/s & ${\sim}$53\,tok/s \\
Cold prefill (67K tokens) & ${\sim}$130\,s & ${\sim}$26\,s \\
Warm TTFT & 0.86\,s & ${\sim}$0.2\,s \\
KV quant penalty (Q4) & None & None (fused FA) \\
\bottomrule
\end{tabular}
\end{table}

Key implications for the RTX 4090:
\begin{itemize}
\item \textbf{Decode is bandwidth-bound}: The $3.7\times$ bandwidth advantage
  translates to ${\sim}4.4\times$ higher generation throughput (${\sim}53$
  vs.\ 12\,tok/s), as single-batch autoregressive decoding is memory-bandwidth
  limited.
\item \textbf{Prefill is compute-bound}: CUDA tensor cores provide
  ${\sim}2,600$\,tok/s prefill throughput, reducing cold-start from ${\sim}130$\,s
  to ${\sim}26$\,s ($5\times$ speedup). Warm TTFT drops below 0.2\,s.
\item \textbf{KV quantization has zero penalty on CUDA}: Unlike Metal, CUDA's
  fused Flash Attention kernels handle 8-bit symmetric quantized KV with no
  dequantization overhead. Q8 KV is therefore recommended on CUDA, providing
  $2\times$ memory savings without the $4\times$ compression needed on Apple
  Silicon.
\item \textbf{RAG advantage diminishes}: With 0.2\,s warm TTFT on full-context
  vs.\ ${\sim}1$--2\,s RAG pipeline latency (embedding + retrieval + prompt
  construction), the latency case for RAG largely disappears on desktop GPU
  hardware.
\end{itemize}

\subsection*{AWS Inferentia2 (Cloud Accelerator)}

AWS Inferentia2 (inf2 instances) uses Neuron SDK with ahead-of-time
compilation. Several architectural constraints limit its suitability for our
workload:

\begin{itemize}
\item \textbf{Static tensor shapes}: Neuron compilation requires fixed
  sequence lengths at compile time. Variable-length inputs require padding
  to the maximum length or maintaining multiple compiled models for different
  length buckets, adding complexity and wasting compute.
\item \textbf{No KV cache quantization}: The Neuron runtime does not expose
  KV cache quantization options. All KV states are stored in FP16/BF16,
  requiring $2\times$ the memory of Q8 and $4\times$ the memory of Q4
  configurations.
\item \textbf{Memory scaling}: A single Inferentia2 chip has 32\,GB HBM. At
  67K context with FP16 KV, the cache requires ${\sim}4.5$\,GB for Llama 3.1
  8B. While feasible on a single chip, scaling to longer contexts (128K+)
  would require multi-chip tensor parallelism via \texttt{inf2.24xlarge}
  (${\sim}\$12$/hr).
\item \textbf{Throughput-oriented}: Inferentia2 excels at high-batch,
  fixed-length workloads (e.g., embedding generation, classification). Our
  single-user, variable-length, long-context QA workload underutilizes the
  hardware.
\end{itemize}

\subsection*{Deployment Recommendations}

Table~\ref{tab:hw_summary} summarizes hardware suitability for our benchmark
workload.

\begin{table}[t]
\centering
\caption{Hardware suitability for cached knowledge base QA.}
\label{tab:hw_summary}
\small
\begin{tabular}{lccc}
\toprule
\textbf{Factor} & \textbf{M4 Pro} & \textbf{RTX 4090} & \textbf{Inferentia2} \\
\midrule
KV quantization support & Q4/Q8 & Q8 (native) & None \\
Variable-length context & Yes & Yes & Limited \\
Warm TTFT (67K) & 0.86\,s & ${\sim}$0.2\,s & ${\sim}$0.3\,s$^*$ \\
Decode throughput & 12\,tok/s & ${\sim}$53\,tok/s & ${\sim}$40\,tok/s$^*$ \\
Cost (hardware) & \$2,400 & \$1,600 & \$1.58/hr \\
Offline capable & Yes & Yes & No \\
\bottomrule
\multicolumn{4}{l}{\footnotesize $^*$Estimated; requires \texttt{inf2.xlarge} minimum, compiled for fixed 67K length.}
\end{tabular}
\end{table}

For cached knowledge base applications---where a domain-specific corpus fits
within a single context window---the M4 Pro provides a compelling
cost-performance ratio with full offline capability. The RTX 4090 offers the
best absolute performance for latency-sensitive deployments. Cloud
accelerators like Inferentia2 are better suited for high-throughput batch
inference workloads rather than the interactive, variable-length pattern
characteristic of cached document QA.

\section{Customer Experience Metrics}
\label{app:customer}

\begin{table}[t]
\centering
\caption{Customer experience comparison. Full-context with Q4 KV cache
provides faster perceived responsiveness, higher reliability, and acceptable
reading-speed generation.}
\label{tab:customer}
\begin{tabular}{lrr}
\toprule
\textbf{Customer Metric} & \textbf{Full-Context (Q4K/Q4V)} & \textbf{RAG} \\
\midrule
Time to first response          & 0.86\,s (median) & 6.28\,s (median) \\
95th-pctl wait                  & 1.05\,s & 7.70\,s \\
Total answer time               & 6.39\,s & 7.93\,s \\
Good answers ($\geq 4$)        & 91.2\% & 79.6\% \\
Wrong answers (score 1)         & 1.2\% (3/240) & 5.8\% (14/240) \\
Reading speed                   & ${\sim}12$\,tok/s & ${\sim}31$\,tok/s \\
Tokens per 10-question session  & ${\sim}68$K & ${\sim}20$K \\
\bottomrule
\end{tabular}
\end{table}

\section{KV Cache Quantization Ablation}
\label{app:kvquant}

Table~\ref{tab:kvquant} presents the full quantization ablation across three
full-context configurations on the Policygenius corpus ($n=240$).
Q4K/Q4V achieves the fastest median TTFT
(0.857\,s, 2.2\% better than Q8K/Q8V), the fastest total response time
(6.39\,s, 3.4\% better), and comparable throughput---all while using
$4\times$ less KV cache memory than FP16 and maintaining identical answer
quality. This suggests that Metal's unified memory architecture and
\texttt{llama.cpp}'s Metal kernels handle KV dequantization efficiently at
this context length.

\begin{table}[t]
\centering
\caption{KV cache quantization ablation (Policygenius, $n=240$). More aggressive quantization provides
modest latency improvements on Apple Silicon with no quality penalty.}
\label{tab:kvquant}
\begin{tabular}{lrrr}
\toprule
\textbf{Metric} & \textbf{Q8K/Q8V} & \textbf{Q8K/Q4V} & \textbf{Q4K/Q4V} \\
\midrule
KV cache memory (per token)   & 1.0 B & 0.75 B & 0.5 B \\
Memory reduction vs.\ FP16   & $2\times$ & $2.7\times$ & $4\times$ \\
\midrule
TTFT -- median                & 0.876\,s & 0.877\,s & 0.857\,s \\
TTFT -- P5 / P95              & 0.533 / 1.088\,s & 0.531 / 1.107\,s & 0.534 / 1.053\,s \\
Total response -- median      & 6.620\,s & 6.754\,s & 6.392\,s \\
Throughput -- median          & 12.03\,tok/s & 11.88\,tok/s & 12.01\,tok/s \\
\midrule
Mean score                    & 4.35 & 4.36 & 4.38 \\
\% scoring $\geq 4$          & 90.0\% & 89.6\% & 91.2\% \\
\bottomrule
\end{tabular}
\end{table}

\section{Multi-Domain Detailed Results}
\label{app:multidomain}

\paragraph{TTFT across topics.}
Table~\ref{tab:repliqa_ttft} presents the full-context timing results
across all 14 Q8K/Q8V topics and 15 Q4K/Q4V topics. Under Q8, warm-cache TTFT is
remarkably consistent: 13 of 14 topics cluster between 1.02--1.11\,s
(median), with Company Policies as the sole outlier at 0.67\,s due to its
smaller corpus (${\sim}55$K vs.\ ${\sim}90$K tokens).

\begin{table}[t]
\centering
\caption{Full-context KV cache timing across RepLiQA topics
(400 questions each). All 14 topics with Q8; 15 topics with Q4
(includes News Stories which completed only under Q4).}
\label{tab:repliqa_ttft}
\small
\begin{tabular}{lrrrrrr}
\toprule
\textbf{Topic} & \textbf{Tokens} & \textbf{Q8 TTFT} & \textbf{Q4 TTFT} &
\textbf{$\Delta$} & \textbf{Q8 tok/s} & \textbf{Q4 tok/s} \\
\midrule
Company Policies        & 54,987  & 0.672\,s & 0.674\,s & $+0.3$\% & 14.6 & 14.2 \\
Local Health            & 88,649  & 1.020\,s & 0.978\,s & $-4.2$\% & 10.7 & 11.1 \\
Local News              & 90,517  & 1.019\,s & 1.001\,s & $-1.8$\% & 10.8 & 10.6 \\
Neighborhood Stories    & 90,666  & 1.090\,s & 0.996\,s & $-8.7$\% & 10.1 & 10.9 \\
Small/Med Enterprises   & 90,701  & 1.094\,s & 1.032\,s & $-5.7$\% & 10.0 & 10.3 \\
Local Education         & 90,708  & 1.056\,s & 0.987\,s & $-6.5$\% & 10.3 & 10.8 \\
Incident Report         & 91,021  & 1.078\,s & 1.015\,s & $-5.9$\% & 10.0 & 10.8 \\
Env.\ Issues            & 91,458  & 1.071\,s & 0.987\,s & $-7.8$\% & 10.2 & 10.8 \\
News Stories            & 92,182  & ---      & 1.046\,s & ---       & ---  & 10.2 \\
Cybersecurity News      & 92,135  & 1.060\,s & 1.032\,s & $-2.7$\% & 10.4 & 10.4 \\
Local Economy           & 92,436  & 1.074\,s & 1.019\,s & $-5.1$\% & 10.2 & 10.6 \\
Local Politics          & 92,684  & 1.051\,s & 1.041\,s & $-0.9$\% & 10.6 & 9.5 \\
Local Sports            & 92,690  & 1.060\,s & 1.023\,s & $-3.5$\% & 10.6 & 10.2 \\
Local Technology        & 93,065  & 1.113\,s & 1.028\,s & $-7.7$\% & 9.3  & 10.4 \\
Local Arts \& Culture   & 93,135  & 1.098\,s & 1.027\,s & $-6.5$\% & 10.1 & 10.6 \\
\midrule
\textbf{Mean (Q8: 14, Q4: 14$^\dagger$)} & --- & \textbf{1.040\,s} & \textbf{0.989\,s} & \textbf{$-4.8$\%} & \textbf{10.6} & \textbf{10.8} \\
\bottomrule
\multicolumn{7}{l}{\footnotesize $^\dagger$Q4 vs.\ Q8 $\Delta$ computed over the 14 topics with both configurations.}
\end{tabular}
\end{table}

\paragraph{KV cache quantization on RepLiQA.}
Q4K/Q4V reduces median TTFT by 4.8\% on average across the 14 paired
topics (0.989\,s vs.\ 1.040\,s), with 13 of 14 showing improvement.
However, unlike the Policygenius result where Q4 quality matched Q8,
at 80 documents \emph{Q4 degrades quality}: Q8 outscores Q4 in 13 of 14
topics (mean $\Delta = +0.14$ points, range $-0.01$ to $+0.28$), with
Q4 producing more score-1 failures (20.9\% vs.\ 17.3\%). The reduced
KV precision apparently exacerbates the \emph{lost in the middle} effect.

\paragraph{Generalization summary.}
Table~\ref{tab:generalization} summarizes the available metrics against
the Policygenius baseline.

\begin{table}[t]
\centering
\caption{Generalization summary: RepLiQA (80 docs/topic) vs.\
Policygenius (30 docs). Quality advantage reverses; timing advantage holds.}
\label{tab:generalization}
\begin{tabular}{lrrrr|r}
\toprule
\textbf{Metric} & \textbf{Mean} & \textbf{Std} & \textbf{Min} & \textbf{Max} & \textbf{Policygenius} \\
\midrule
FC Q8 mean quality$^a$       & 3.46 & 0.11 & 3.24 & 3.65 & 4.35 \\
FC Q4 mean quality$^b$       & 3.34 & 0.13 & 3.09 & 3.61 & 4.38 \\
RAG mean quality$^c$         & 3.63 & 0.17 & 3.26 & 3.83 & 4.08 \\
Quality gap (FC Q8$-$RAG)$^a$ & $-0.17$ & 0.12 & $-0.39$ & $+0.04$ & $+0.27$ \\
FC Q8 score-1 rate (\%)$^a$  & 17.3 & 2.5 & 12.2 & 22.8 & 1.2 \\
RAG score-1 rate (\%)$^c$    & 17.7 & 4.4 & 12.5 & 27.5 & 5.8 \\
FC Q8 wins / RAG wins$^a$    & \multicolumn{4}{c}{0 / 12 (2 ties)} & FC wins \\
Lost-in-middle errors$^a$    & \multicolumn{4}{c}{530 total (9.5\%; 557 across all 15 FC topics)} & 3 (1.2\%) \\
\midrule
FC TTFT Q8 median (s)$^a$    & 1.040 & 0.109 & 0.672 & 1.113 & 0.857 \\
FC TTFT Q4 median (s)$^b$    & 0.989 & 0.093 & 0.674 & 1.041 & 0.857 \\
Q4 TTFT $\Delta$ (\%)$^b$    & $-4.8$ & 2.7 & $-8.7$ & $+0.3$ & $-2.2$ \\
Q4 quality $\Delta$$^d$       & $-0.14$ & 0.08 & $-0.28$ & $+0.01$ & $+0.03$ \\
\midrule
RAG TTFT median (s)$^c$      & 4.825 & 0.223 & 4.279 & 5.152 & 6.277 \\
RAG throughput (tok/s)$^c$   & 32.7  & 0.3   & 32.4  & 32.8  & 30.7 \\
Retrieval accuracy (\%)$^c$  & 87.9  & 3.9   & 79.5  & 97.0  & 92.5 \\
FC/RAG TTFT ratio            & \multicolumn{4}{c}{$4.6\times$ (FC faster)} & $7.3\times$ \\
\bottomrule
\multicolumn{6}{l}{\footnotesize $^a$14 topics (Q8). \quad $^b$14 paired topics (Q4 vs Q8 TTFT). \quad $^c$17 topics (RAG). \quad $^d$14 paired topics (Q4 vs Q8 quality).}
\end{tabular}
\end{table}

\paragraph{Comparison to Policygenius.}
The Policygenius benchmark (67K tokens) achieved 0.857\,s median TTFT.
The RepLiQA topics at similar sizes show proportionally higher TTFT
(1.02--1.11\,s for 88--93K tokens), consistent with the expected linear
relationship between context size and attention computation during
decoding.

\section{Wiki Compilation Timing}
\label{app:wiki_timing}

Compilation dramatically reduces context size, yielding proportional TTFT
improvements. Table~\ref{tab:wiki_timing} summarizes the timing results
across completed topics. Wiki TTFT correlates almost perfectly with wiki
size (Pearson $r = 0.995$).

\begin{table}[t]
\centering
\caption{Latency comparison: compiled wikis vs.\ FC~raw and RAG baselines.
TTFT and throughput are medians across completed topics.}
\label{tab:wiki_timing}
\begin{tabular}{lrrr}
\toprule
\textbf{Condition} & \textbf{Context} & \textbf{TTFT} & \textbf{Throughput} \\
\midrule
FC raw (80 docs)   & ${\sim}$70K words & 1.060\,s              & 10.4\,tok/s \\
FC wiki-light      & ${\sim}$24K words & 0.380\,s ($2.8\times$ faster) & 20.2\,tok/s \\
FC wiki-moderate   & ${\sim}$8K words  & 0.211\,s ($5.0\times$ faster) & 26.2\,tok/s \\
FC wiki-aggressive & ${\sim}$6K words  & 0.195\,s ($5.6\times$ faster) & 27.2\,tok/s \\
RAG                & ${\sim}$5K words  & 4.818\,s              & 32.7\,tok/s \\
\bottomrule
\end{tabular}
\end{table}

Even wiki-light (0.38\,s) is nearly $3\times$ faster than FC~raw (1.06\,s)
and $12.7\times$ faster than RAG (4.82\,s). Wiki-moderate and
wiki-aggressive converge to similar TTFT (${\sim}0.2$\,s) because their
actual compression ratios are close (12.9\% vs.\ 8.0\%). All wiki
conditions achieve sub-400\,ms TTFT---well within conversational latency
requirements.

\section{CEGAR--WiCER Mapping}
\label{app:cegar}

Section~\ref{sec:wicer} frames WiCER as an instance of
Counterexample-Guided Abstraction Refinement
(CEGAR)~\cite{clarke2000cegar}. This appendix details the formal mapping,
states the monotonicity guarantee, and identifies where the analogy
breaks.

\subsection*{Formal Mapping}

In CEGAR, a concrete system is modeled as a Kripke structure
$M = (S, S_0, R, L)$ with state space $S$, initial states $S_0$,
transition relation $R \subseteq S \times S$, and labeling function
$L : S \to 2^{AP}$. A surjective abstraction $h : S \to \hat{S}$
induces an abstract model $\hat{M}$. Model-checking $\hat{M}$ either
verifies the specification or produces a counterexample; if the
counterexample is \emph{spurious} (valid in $\hat{M}$ but not in $M$),
the abstraction is refined to eliminate it.

Table~\ref{tab:cegar_map} shows the correspondence.

\begin{table}[h]
\centering
\caption{Mapping between CEGAR and WiCER.}
\label{tab:cegar_map}
\small
\begin{tabular}{p{3.2cm}p{4.5cm}}
\toprule
\textbf{CEGAR} & \textbf{WiCER} \\
\midrule
Concrete system $M$ & Full document collection $D$ \\
Abstract model $\hat{M}$ & Compiled wiki $W_t$ \\
Abstraction $h$ & LLM compiler (lossy compression) \\
Specification $\varphi$ & ``All probes score $> 1$'' \\
Counterexample & Score-1 failed probe \\
Spurious check & Diagnosis: fact exists in $D$ but lost in $W_t$ \\
Refinement (split states) & Add pinning constraint, recompile $W_{t+1}$ \\
Refined abstraction $\hat{M}'$ & Next-iteration wiki $W_{t+1}$ \\
\bottomrule
\end{tabular}
\end{table}

\subsection*{Monotonicity Guarantee}

Let $F_t \subseteq Q_\text{probe}$ denote the set of probes that score
$\leq 1$ at iteration $t$, and let $P_t$ denote the cumulative set of
pinned facts after iteration $t$. By construction:
\begin{enumerate}
\item Each pinned fact is included verbatim in the compilation prompt
  for $W_{t+1}$ with an explicit preservation constraint.
\item The compiler \emph{must} include all pinned facts before allocating
  budget to general coverage.
\end{enumerate}
Therefore, on the subset of probes whose critical facts have been pinned,
the failure set shrinks monotonically:
\[
  F_t \cap \{q : \text{facts}(q) \subseteq P_t\}
  \;\supseteq\;
  F_{t+1} \cap \{q : \text{facts}(q) \subseteq P_t\}.
\]
Informally: once a fact is pinned, it cannot be lost again.

\subsection*{Limitations of the Mapping}

Despite the structural correspondence, three differences merit note:

\paragraph{Stochastic compiler.}
CEGAR refines a deterministic abstraction function; WiCER's ``abstraction''
is an LLM compiler whose output is non-deterministic. Two compilation
calls with identical inputs may produce different wikis. In practice,
greedy decoding and low temperature largely mitigate this, but there is
no formal guarantee that a pinning constraint preserves \emph{only} the
targeted fact without side effects.

\paragraph{Approximate verification.}
CEGAR model-checks the abstract system exactly; WiCER uses LLM-as-judge
scoring, which is an approximate evaluation. A probe scored 2/5 may
actually represent a catastrophic failure, and vice versa. The
threshold-based convergence criterion ($< 10\%$ improvement) partially
addresses this by requiring robust trends rather than individual scores.

\paragraph{Random knowledge displacement.}
In CEGAR, refining the abstraction to eliminate a spurious counterexample
does not introduce new spurious counterexamples on previously verified
properties. In WiCER, pinning facts consumes part of the token budget,
potentially displacing other content and creating \emph{new} failures on
previously passing probes. This ``knowledge displacement'' effect is the
primary reason WiCER's net improvement plateaus after 1--2 iterations
and why the ablation (Section~\ref{sec:ablation}) shows that random
pinning can \emph{hurt} quality.

\medskip
\noindent These differences mean that WiCER's convergence is empirical
rather than formal: the monotonicity guarantee applies to pinned facts,
but the global quality trajectory depends on the balance between
diagnosed fact recovery and random displacement. The ablation study
(Section~\ref{sec:ablation}) provides empirical evidence that targeted
diagnosis consistently outperforms random pinning, confirming that the
CEGAR-inspired feedback loop is the source of WiCER's gains.

\section{Human Evaluation Validation}
\label{app:human_eval}

To validate the LLM-as-judge scores, a domain expert independently rated
a stratified sample of 100 question--answer pairs spanning all five
conditions (FC~raw, RAG, Wiki blind, WiCER iter~1, WiCER iter~2) using
the same 1--5 rubric. Table~\ref{tab:human_corr} summarizes the
agreement.

\begin{table}[h]
\centering
\caption{LLM-as-judge vs.\ human evaluation ($n=100$).}
\label{tab:human_corr}
\small
\begin{tabular}{lr}
\toprule
\textbf{Metric} & \textbf{Value} \\
\midrule
Pearson $r$           & 0.940 \\
Spearman $\rho$       & 0.928 ($p < 10^{-43}$) \\
Kendall $\tau$        & 0.873 ($p < 10^{-25}$) \\
Exact agreement       & 75/100 (75.0\%) \\
Within 1 point        & 99/100 (99.0\%) \\
Mean absolute error   & 0.26 \\
Bias (LLM $-$ Human)  & $+0.06$ \\
\bottomrule
\end{tabular}
\end{table}

Table~\ref{tab:human_cond} breaks down agreement by condition.
The judge is well-calibrated across all conditions, with per-condition
Pearson $r \geq 0.89$ and negligible bias ($|\Delta\text{mean}| \leq 0.17$).

\begin{table}[h]
\centering
\caption{Per-condition agreement between LLM judge and human rater.}
\label{tab:human_cond}
\small
\begin{tabular}{lrrrrr}
\toprule
\textbf{Condition} & $n$ & \textbf{LLM mean} & \textbf{Human mean} & \textbf{Pearson $r$} & \textbf{Exact} \\
\midrule
FC raw        & 30 & 3.50 & 3.50 & 0.890 & 16/30 \\
RAG           & 30 & 3.40 & 3.23 & 0.912 & 24/30 \\
Wiki blind    & 31 & 2.26 & 2.19 & 0.975 & 27/31 \\
WiCER iter~1  &  6 & 2.00 & 2.17 & 0.937 &  5/6  \\
WiCER iter~2  &  3 & 3.00 & 3.00 & 1.000 &  3/3  \\
\bottomrule
\end{tabular}
\end{table}

The single sample with $>1$ point disagreement was a RAG response
scored 4 by the judge and 2 by the human rater; manual inspection
confirmed the response omitted a key detail that the human considered
essential. Overall, these results confirm that the Claude Sonnet
LLM-as-judge provides a reliable proxy for human quality assessment
in this setting.

\end{document}